\documentclass[10pt,twocolumn,letterpaper]{article}

\usepackage{wacv}
\usepackage{times}
\usepackage{epsfig}
\usepackage{graphicx}
\usepackage{amsmath}
\usepackage{amssymb}
\usepackage{bbm}
\DeclareMathOperator*{\argmax}{argmax}
\usepackage{floatrow} 
\usepackage{makecell}

\def\wacvPaperID{655} 

\wacvfinalcopy 

\ifwacvfinal
\fi

\ifwacvfinal
\usepackage[breaklinks=true,bookmarks=false]{hyperref}
\else
\usepackage[pagebackref=true,breaklinks=true,colorlinks,bookmarks=false]{hyperref}
\fi

\begin{document}

\title{Hierarchical Modeling for Task Recognition and Action Segmentation in Weakly-Labeled Instructional Videos}

\author{Reza Ghoddoosian \hspace{2cm} Saif Sayed \hspace{2cm} Vassilis Athitsos \\
Vision-Learning-Mining Lab, University of Texas at Arlington\\
{\tt\small \{reza.ghoddoosian, saififtekar.sayed\}@mavs.uta.edu, athitsos@uta.edu}}

\maketitle
\thispagestyle{empty}

\begin{abstract}
   This paper \footnote{ https://github.com/rezaghoddoosian/Hierarchical-Task-Modeling} focuses on task recognition and action segmentation in weakly-labeled instructional videos, where only the ordered sequence of video-level actions is available during training. We propose a two-stream framework, which exploits semantic and temporal hierarchies to recognize top-level tasks in instructional videos. Further, we present a novel top-down weakly-supervised action segmentation approach, where the predicted task is used to constrain the inference of fine-grained action sequences. Experimental results on the popular Breakfast and Cooking 2 datasets show that our two-stream hierarchical task modeling significantly outperforms existing methods in top-level task recognition for all datasets and metrics. Additionally, using our task recognition framework in the proposed top-down action segmentation approach consistently improves the state of the art, while also reducing segmentation inference time by 80-90 percent.  
\end{abstract}

\section{Introduction}

Millions of people watch instructional videos online every day, to learn to perform tasks such as cooking or changing a car tire. Also, new models of assistant robots \cite{kitchen_robot} can learn from such videos how to assist humans in their daily lives. Hence, there has been extensive research in recent years on automated understanding of the top-level tasks and their sub-actions in such videos \cite{embedding,Unsupervised_Elhamifar,Cooking,COIN}.

From a theoretical point of view, instructional videos can be seen as videos illustrating hierarchical activities. Each instructional video illustrates a single top-level activity, for which we use the term ``task'' throughout this paper. Examples of such video-level tasks are ``making coffee'' or ``cooking eggs''. Each video-level task is composed of a sequence of lower level activities, such as ``pouring milk'' or ``adding sugar''. Throughout the paper, we will  refer to such lower-level activities using the term ``action''.  Consequently,  using this terminology,  each instructional video illustrates a task  that consists of a sequence of actions.

For instructional videos, and hierarchical activity videos  in general,  we would like  to have automated systems that both recognize the overall task and also understand what lower-level actions take place, and when those actions start and end. Fully-supervised training would require not only  annotating the top level task, but also marking the start and end frame of each lower-level action. With the ever-growing size of instructional video datasets, manually annotating such start and end frames can quickly become a bottleneck. To address this issue, {\em weakly-supervised action segmentation} methods require, as training data, only the sequence of actions that takes place at each video, and no start/end frame information for those actions \cite{D3TW,ActPrototype,TCFPN,CDFL,NNViterbi}. 

\begin{figure}[t]
\begin{center}

   \includegraphics[width=1\linewidth]{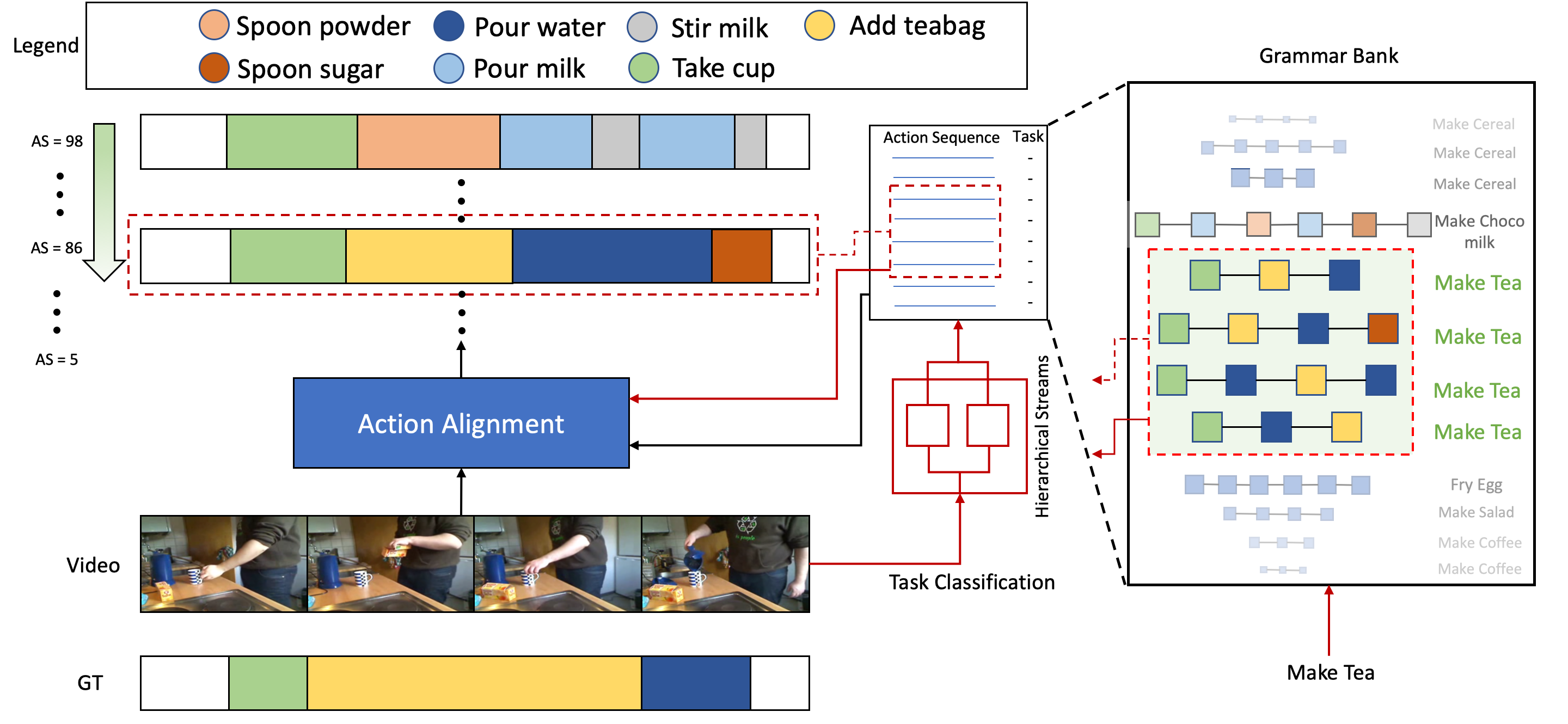}
\end{center}
   \caption{The proposed segmentation approach. Initially, the weak understanding of duration leads to the alignment of irrelevant actions with an Alignment Score (AS) of 98. However, a prior prediction of the task allows the alignment module to infer the correct sequence of actions (outlines by red) despite the lower AS (86).}
\label{fig:segmentation}
\end{figure}

Our goal in this paper is to jointly address the problems of top-level task recognition and lower-level action segmentation, in the weakly supervised setting (given sequences of actions, not given start/end frames). The main novelty is a method for top-level task recognition that uses, in parallel, two different hierarchical decompositions of the problem. One module models the semantic hierarchy between the top-level task and lower-level attributes. These attributes correspond to either the set of actions, or the set of the object/verb components of those actions, e.g., ``take'' and ``cup'' in the action ``take cup''. This module jointly learns to identify the presence of attributes in the video and to recognize the top-level task based on the estimated attributes.

A  parallel second stream models the temporal hierarchy between the entire video and equal-duration subdivisions of the video. Tasks are usually performed in a relative order of stages, and some stages  are particularly useful for distinguishing tasks from each other. For example, preparing tea typically involves three stages: taking a cup, adding a tea bag, and pouring water from the kettle. The first and last stages are visually similar with corresponding  stages of the ``preparing coffee'' task. The temporal hierarchy module can capture the importance of adding the tea bag in distinguishing the ``preparing tea''  task from the ``preparing coffee'' task. This module learns the relation between stages and their importance in classifying the video task.

We also propose a novel top-down approach for action segmentation (i.e., frame-level action labeling), that combines our task recognition method with existing weakly-supervised segmentation methods. In this approach (Fig.\ref{fig:segmentation}), the video-level task is estimated first, and is subsequently used to constrain the search space for action segmentation.
In summary, the contributions of this paper are these:

1) We introduce a two-stream framework that exploits both semantic and temporal hierarchies to recognize tasks in weakly-labeled instructional videos. 

2) We provide specific, non-trivial implementations of these two streams. Our ablation studies demonstrate that our implementation choices have a significant impact on performance. A highlight of such an implementation choice is using TF-IDF weights to model the discriminative power of each attribute for each task (see Section \ref{meth:sh}). 

3) We present a novel top-down approach for weakly-supervised action segmentation, where the video-level task is used to constrain the segmentation output. 

4) We present results on two benchmark datasets: Breakfast \cite{Breakfast} and MPII Cooking 2 \cite{Cooking}. In top-level task recognition, our method significantly outperforms the state of the art on both datasets for all metrics. For weakly supervised  action segmentation (frame-level labeling), applying the proposed top-down approach on top of existing methods \cite{CDFL,NNViterbi} again leads to state of the art results, and also cuts the inference time by 80-90 percent.







\section{Related Work}
\textbf{Instructional Video Analysis.} In recent years, untrimmed instructional videos have been studied in areas like video retrieval \cite{embedding,representation,retrieval}, quality assessment \cite{ranking1,assessment}, future action planning \cite{chang2019procedure}, and key-step segmentation \cite{Unsupervised_Elhamifar,TEC,set-constrained,actionset,Cooking,COIN,zhou2017towards}. Fully-supervised action segmentation methods \cite{fullysupervised5,fullysupervised2,kuehneWACVend,fullysupervised3,RichardStatLang,fullysupervised1,fullysupervised4} learn to identify action segments in the presence of frame-level ground-truth. For example, \cite{COIN} use a bottom-up technique to aggregate initial action proposal scores to classify the top-level video task, before modifying its preliminary frame labels in a fully supervised way. Also, \cite{Cooking} analyze a host of holistic and regional features to train shared low-level classifiers to recognize tasks and detect fine-grained actions. 

Recently, unsupervised learning of instructional videos has seen increased attention \cite{Unsupervised_Elhamifar,summarization,TEC,sener2018unsupervised}. In \cite{TEC}, an unsupervised approach performs video segmentation and task clustering through learned feature embeddings. In \cite{Unsupervised_Elhamifar}, a network is trained using only video task labels, for unsupervised discovery of procedure steps and task recognition. 

The above-mentioned methods are either fully supervised or unsupervised, and thus they are not direct competitors for our method, which uses weak labels.

In the scope of activity recognition, most works \cite{x3d,stm,nlb} study short-range or trimmed videos. Our work is closest to \cite{timeception,videograph,rhyrnn}, where the focus is recognizing minutes-long activities. However, unlike them, our paper is on instructional videos, and on how recognition can aid segmentation, so it relies on hierarchical activity labels (top-level task, lower-level attributes as targets for segmentation). 

\textbf{Weakly-Supervised Key-Step Localization.}
In the context of weakly-labeled instructional videos, many methods \cite{narration1,narration2,whats_cooking,crosstask} are trained under the supervision of narration and subtitle. Directly relevant to our work are \cite{D3TW,ActPrototype,TCFPN,ECTC,CDFL,Fine2Coarse,NNViterbi}, where, as in our method, only the sequence of actions is known for each training video. In particular \cite{CDFL,NNViterbi} deploy a factorized probabilistic model to tackle the segmentation problem using dynamic programming. Also, \cite{D3TW} formulate a differential dynamic programming framework for end-to-end training of their model. 

Recent weakly-supervised segmentation methods \cite{D3TW,ActPrototype,CDFL,NNViterbi} are formulated to identify the action taking place at each frame, and not the top-level video task. At the same time, the output of these methods implicitly specifies the top-level task, because only one task is compatible with the detected sequence of actions. We use these implicit task predictions of \cite{CDFL,NNViterbi} to compare those methods to ours on task recognition accuracy. In contrast to these bottom-up approaches (going from actions to task), our method explicitly learns to classify video-level tasks, and this classification is used in a top-down fashion (from task to actions) to constrain the detected action sequence. 

We should also mention the methods in \cite{3C,detection1,WTALC,TSN}, which perform weakly-supervised action \textit{detection}. These methods identify and localize occurrences of, typically, a single action in the input video. For completeness, we evaluate extensions of these methods to task classification.

\section{Hierarchical Task Modeling Method} \label{meth:classification}
In this section, we present an overview of our two-stream hierarchical task modeling. Full details of our implementation choices and architecture are provided in Sec. \ref{sec:method_details}.

As our formulation uses many terms and symbols, the supplementary material provides a glossary of terms and a table of all symbols we use.

\subsection{Method Overview}\label{meth:overview}

\textbf{Problem Definition.}
The training set $\mathbb{V}=\{\mathbf{v}_i\}_{i=1}^N$ consists of N videos $\mathbf{v}_i$. From each $\mathbf{v}_i$ we extract a feature vector $\mathbf{x}_i\in \mathbb{R}^{F\times T_i}$, that consists of $T_i$ frames of $F$-dimensional features. We denote by $\mathbb{C}=\{c_{i}\}_{i=1}^{|\mathbb{C}|}$ the set of all top-level task labels, and by $\mathbb{A}=\{a_{j}\}_{j=1}^{|\mathbb{A}|}$ the set of all lower-level attribute labels. As an implementation choice, these attributes can be the set of actions in the dataset, or the set of verb/object components of those actions. Each video $\mathbf{v}_i$ is labeled by a task $c_i \in \mathbb{C}$, and also by a set $\mathbb{A}_i \subseteq \mathbb{A}$ of $M_i$ attributes, so that $\mathbb{A}_i=\{a_{i,j}\}_{j=1}^{M_i}$. At test time, given an input video, the system estimates the top-level task.

\textbf{Semantic Hierarchy Stream (SHS).}
To recognize the task, one approach is to directly estimate $p(c_i|\mathbf{x}_i)$. However, this approach is prone to overfitting when the number of video samples per task is limited. As attributes can be shared among tasks, the average number of training videos per attribute is typically greater than the average number of videos per task. Using attribute information also helps the model learn similarities and differences of spatio-temporal patterns in different tasks. 

\begin{figure}[t]
\begin{center}
   \includegraphics[width=0.75\linewidth]{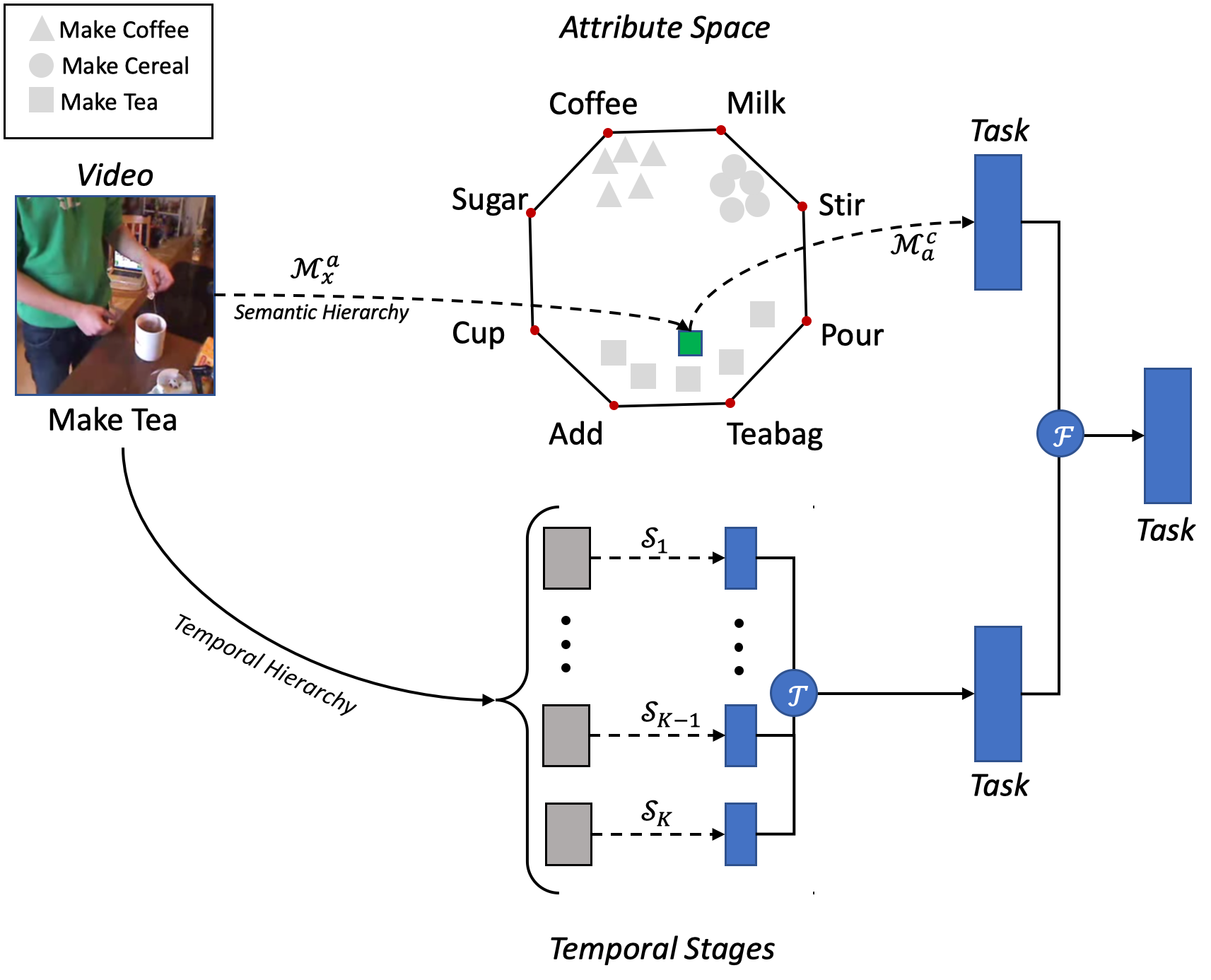}
\end{center}
   \caption{An overview of how our recognition model exploits the semantic and temporal hierarchies of tasks. The attribute representation of videos are formed by their discriminative attributes.}
\label{fig:overview}
\end{figure}

Thus, we model task recognition as $p(c_i|\psi_i^a).p(\psi_i^a|\mathbf{x}_i)$, where $\mathbf{\psi}_i^a$ is an intermediate vector of attribute scores that is computed for each $\mathbf{x}_i$. The system learns a mapping function $\mathcal{M}_x^a:\mathbb{R}^{F\times T_i} \rightarrow \mathbb{R}^{|\mathbb{A}|}$, that maps each vector $\mathbf{x}_i$ to attribute score vector $\mathbf{\psi}_i^a$. It also learns a function $\mathcal{M}_a^c:\mathbb{R}^{|\mathbb{A}|}\rightarrow \mathbb{R}^{|\mathbb{C}|}$, that maps each attribute score vector $\psi_i^a$ to a task score vector $\psi_i^c$ (Fig.\ref{fig:overview}).


\textbf{Temporal Hierarchy Stream (THS).}
Tasks in instructional videos are usually performed in a relative order of steps. Understanding the task-discriminative stages of a video is essential in distinguishing tasks that share similar-looking actions. Thus, we divide each video into $K$ stages of equal duration, and train a classifier $\mathcal{S}_\kappa:\mathbb{R}^{F\times\frac{T_i}{K}}\rightarrow \mathbb{R}^{|\mathbb{C}|}$ for each stage $\kappa$. The system also learns an aggregation function $\mathcal{T}:\mathbb{R}^{K|\mathbb{C}|}\rightarrow \mathbb{R}^{|\mathbb{C}|}$, that maps stage-wise predictions to classification scores $\vartheta_{i,total}$ of the entire video.


\textbf{Stream Fusion Module.} 
In the end, we fuse the predictions of the SHS and THS streams to output the final task prediction scores $f_i^c$ of the entire model. A high-level diagram of the overall network is shown on Fig.\ref{fig:overview}. The network is optimized using a loss function for the fusion module, as well as separate loss functions for the SHS and THS streams.

\subsection{Detailed Architecture}\label{sec:method_details}
\begin{figure*}
    \includegraphics[width=0.9\linewidth]{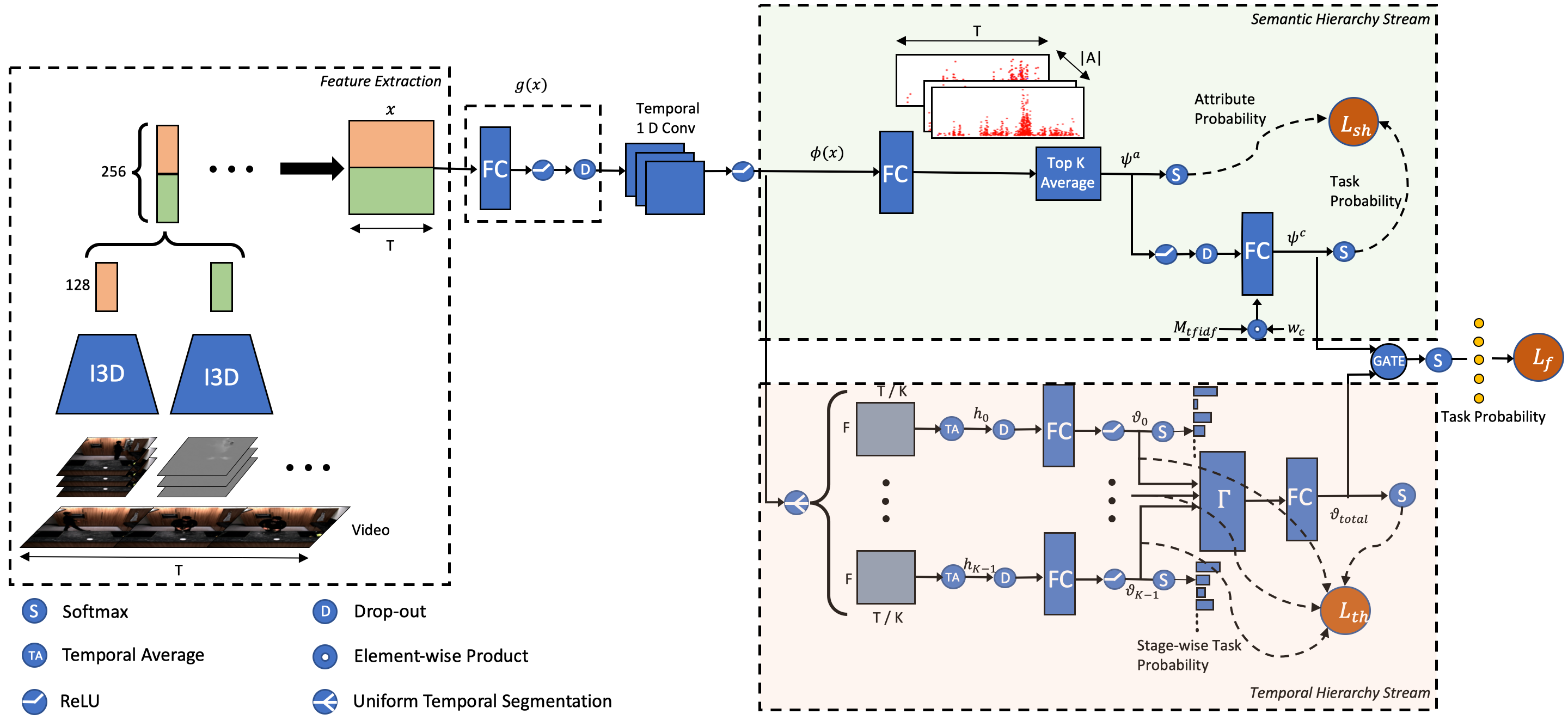}
    \caption{Our two-stream model architecture for task classification using RGB and flow frames as input. The semantic hierarchy loss $\mathcal{L}_{sh}$ ensures task classification after clustering videos based on their shared weighted attributes through the TF-IDF mask $\mathbf{M}_{tfidf}$. The THS stream learns to aggregate stage-wise task predictions by $\mathcal{L}_{th}$, and a third loss ($\mathcal{L}_{f}$) optimizes the fused results of both streams. }
    \label{fig:architecture}
\end{figure*}

In this section we explain in detail the architecture of our two-stream hierarchical model (Fig.\ref{fig:architecture}), and we derive the three proposed loss functions.

\subsubsection{Feature Extraction}
Video task recognition is highly dependent not only on motion patterns, but also on object appearance. Ignoring object appearance can lead to misclassifications when the motion patterns of two tasks are very similar, e.g., making coffee and making tea. Hence, instead of the mostly motion-based iDT features\cite{iDT} used in \cite{D3TW,CDFL,NNViterbi}, we adopt the I3D network, pre-trained on the Kinetics dataset\cite{I3D}. I3D extracts, for each frame, 1024-dimensional feature vectors respectively from the RGB and optical flow channels. We use PCA separately on RGB and flow features, to reduce the dimensions from 1024 to 128. 

The 256-dimensional concatenated RGB and optical flow features of each frame are stored in video-level feature vector $\mathbf{x}_i\in \mathbb{R}^{256\times T_i}$, where $T_i$ is the total number of frames in video $\mathbf{v}_i$. In principle, any spatio-temporal network can be used instead of I3D. In the supplementary material, we show that I3D outperforms iDT for task recognition.    

\subsubsection{Semantic Hierarchy Loss}\label{meth:sh}
In order to obtain a data-specific representation of video-level feature vector $\mathbf{x}_i$, we pass each frame-level subvector of $\mathbf{x}_i$ through a fully-connected layer $g$ with bias and output dimension of 256, then apply temporal convolution to the output of $g$. Using such 1D temporal convolutions with a set of $F$ learnable kernels $\mathbf{k}_{\phi}\in \mathbb{R}^{L \times 256}$ of size $L$ , $\mathbf{x}_i$ is eventually mapped to a $F$-dimensional feature encoding $\mathbb{\phi}(\mathbf{x}_i)\in \mathbb{R}^{F\times T_i}$.

Let $|\mathbb{A}|$ be the number of unique attributes in the dataset. We pass each frame-level subvector of $\mathbb{\phi}(\mathbf{x}_i)$ through a fully connected layer with bias and output dimension of $|\mathbb{A}|$ to obtain $\Psi_i^a\in \mathbb{R}^{|\mathbb{A}|\times T_i}$, which is the sequence of attribute scores for each of the $T_i$ frames. $\Psi_i^a$ is also known as {\em temporal class activation map} (T-CAM)\cite{TCAM}.

Similar to \cite{WTALC,WTALC-graph}, we compute a video-level attribute score vector $\psi^a_i$ by average-pooling the highest $k_i=\lfloor\frac{T_i}{s}\rfloor$ T-CAM scores of each attribute separately over time, where $s$ is a hyperparameter:
\begin{small}
\begin{equation}
    \psi_{i}^a[j]=\frac{1}{k_i}\sum_{i=0}^{k_i-1}\mathrm{top_k}\{\Psi_i^a[j,:]\}
\end{equation}
\end{small}
Intuitively, these selected $k_i$ scores highlight the most important parts of a task in video $i$. 

To denote the set $\mathbb{A}_i$ of attributes present in video $\mathbf{v}_i$, we define a multihot ground-truth attribute vector $\vec{a}_{i}\in \{0,1\}^{|\mathbb{A}|}$, where for every $j\in \{0,1,...,|\mathbb{A}|-1\}$, $\vec{a}_{i,j}=1$ if $a_{j} \in \mathbb{A}_i$, otherwise $\vec{a}_{i,j}=0$. However, this representation fails to capture the fact that different attributes have different levels of relevance for recognizing each task. For example, attributes ``take out'' and ``open'' are present in most videos, and thus not discriminative. As a second example, for the ``preparing avocado'' task, ``avocado'' is a more informative attribute compared to ``knife''.

 Inspired by text retrieval methods \cite{tfidf1,tfidf2}, we compute the TF-IDF weight matrix $\mathbf{W}_{tfidf}\in\mathbb{R}^{|\mathbb{A}|\times |\mathbb{C}|}$, so that $\mathbf{W}_{tfidf}(j,\tau)$ captures the importance of attribute $j$ for task $\tau$. Initially, we formulate $\mathrm{TF}\in \mathbb{R}^{|\mathbb{A}|\times |\mathbb{C}|}$ and $\mathrm{IDF}\in \mathbb{R}^{|\mathbb{A}|}$ as follows:
\begin{small}
\begin{align}
\mathrm{TF}(j,\tau)={}&\frac{\sum_{i=0}^{N-1} \vec{a}_{i,j}\cdot \mathbbm{1}(\tau=c_i)}{ \sum_{i=0}^{N-1}\mathbbm{1}(\tau=c_i)}\\
\mathrm{IDF}(j)={}&\mathrm{log}(\frac{|\mathbb{C}|}{| \{\tau\in \mathbb{C} | \mathrm{TF}(j,\tau)>0\}|})
\label{}
\end{align}
\end{small}
where $\mathbbm{1}()$ denotes the indicator function and $\mathrm{TF}(j,\tau)$ and $\mathrm{IDF}(j)$  are, respectively, the percentage of times attribute $a_j$ is present in videos of task $\tau\in \mathbb{C}$, and the log inverse of percentage of all tasks that entail attribute $j$ in at least one of their videos. We then define the elements $\mathbf{W}_{tfidf}(j,\tau)$ of the TF-IDF weight matrix  as:
\begin{small}
\begin{align}
\mathbf{W}_{tfidf}(j,\tau)=\frac{\mathrm{TF}(j,\tau) \cdot \mathrm{IDF}(j)}{\epsilon+\sum_{k=0}^{|\mathbb{A}|-1}\mathrm{TF}(k,\tau) \cdot \mathrm{IDF}(k)}
\label{}
\end{align}
\end{small}
with $\epsilon$ set to a very small value to avoid division by zero. 

Using these TF-IDF weights, we intorduce the TF-IDF-weighted attribute ground-truth vector $\vec{a^w}_i\in \mathbb{R}^{|\mathbb{A}|}$ as:
\begin{small}
\begin{align}
\vec{a^w}_{i,j}=\frac{\vec{a}_{i,j}\cdot \mathbf{W}_{tfidf}(j,c_i)}{\sum_{k=0}^{|\mathbb{A}|-1}\vec{a}_{i,k}\cdot \mathbf{W}_{tfidf}(k,c_i)}  
\label{}
\end{align}
\end{small}

We also define a TF-IDF mask $\mathbf{M}_{tfidf}\in \mathbb{R}^{|\mathbb{A}|\times |\mathbb{C}|}$, where $\mathbf{M}_{tfidf}(j,\tau)$ is 1 if the corresponding TF-IDF weight $\mathbf{W}_{tfidf}(j,\tau)$ is nonzero, otherwise it is 0. We use the TF-IDF mask to form a mapping from attribute score vectors $\psi_i^a$ to task probability scores $\hat{\psi_i^c}\in \mathbb{R}^{|\mathbb{C}|}$, as:
\begin{small}
\begin{align}
\hat{\psi_i}^{c}=\mathfrak{s}[(\mathbf{w}_c \odot \mathbf{M}_{tfidf}^T)\mathrm{ReLU}(\psi_i^a)]
\label{}
\end{align}
\end{small}
In the above, $\mathfrak{s}[]$ and $\odot$ mean the softmax and element-wise product operations respectively, and $\mathbf{w}_c\in\mathbb{R}^{|\mathbb{C}|\times |\mathbb{A}|}$ are weights to be learned. Using the TF-IDF mask allows the model to focus only on relevant attributes for each task.

Let $\vec{c}_i$ be the one-hot task ground-truth vector and $\hat{\psi^a}=\mathfrak{s}[\psi^a]$. The semantic hierarchy loss $\mathcal{L}_{sh}$ is then defined as:
\begin{small}
\begin{align}
\mathcal{L}_{sh}=-\lambda \mathbb{E}[\vec{a^w}_i^T log(\hat{\psi_i^a})]- (1-\lambda) \mathbb{E}[\vec{c}_i^T log(\hat{\psi_i}^{c})]
\label{eq:sh_loss}
\end{align}
\end{small}
$\mathbb{E}$ denotes ``expected value'', and $\lambda$ is a design parameter that decides how fast each term is trained comparatively.

\subsubsection{Temporal Hierarchy Loss}
We model the temporal hierarchy by dividing each video into $K$ stages of equal duration $d$, and training a classifier for each stage. Formally, given the frame-level feature encoding $\phi(\mathbf{x}_i)$ of video $i$, we define $h_{i,\kappa}\in \mathbb{R}^{F}$ as the feature summary of the $\kappa$-th stage and produce unnormalized task scores (logits) $\vartheta_{i,\kappa}\in \mathbb{R}^{|\mathbb{C}|}$: 
\begin{small}
\begin{align}
h_{i,\kappa}=&{} \frac{\sum_{t=\kappa d}^{[(\kappa+1) d]-1}\phi(\mathbf{x}_i)[:,t] }{d}\\
\vartheta_{i,\kappa}=&{}\mathbf{w}_{\kappa}h_{i,\kappa} + b_{\kappa}
\label{}
\end{align}
\end{small}
where $\mathbf{w}_{\kappa}\in \mathbb{R}^{|\mathbb{C}|\times F}$ and $b_{\kappa}\in \mathbb{R}^{|\mathbb{C}|}$ are parameters of each stage. During the training process, for each stage, the loss function $\mathcal{L}_{\kappa}=\mathbb{E}[\vec{c}_i^T log(\hat{\vartheta}_{i,\kappa})]$ is defined on the softmax of the stage-wise task prediction logits $\hat{\vartheta}_{i,\kappa}$.

\textbf{Stage Aggregation Function.} As mentioned earlier, certain stages of a task are more discriminative than others. We define the auxiliary function $\Gamma([\vartheta_{0}\ \vartheta_{1}\ ...\ \vartheta_{K-1}])$, that maps stage-level task score vectors to a video-level task score vector. While training, $\Gamma(\ )$ randomly masks out one of its $K$ input prediction vectors $\vartheta_{\kappa}$ entirely and multiplies the rest of the input predictions by $\frac{K}{K-1}$. $\Gamma(\ )$ acts similar to the spatial drop out operation by promoting independence between predictions of each stage, to avoid overfitting to a single stage. We form our stage aggregation function using $\Gamma(\ )$ and aggregation parameters $\mathbf{w}_{total}\in \mathbb{R}^{K |\mathbb{C}|\times |\mathbb{C}|}$ to produce video-level task probability values $\hat{\vartheta}_{i,total}\in \mathbb{R}^{|\mathbb{C}|}$:
\begin{small}
\begin{align}
\hat{\vartheta}_{i,total}^T=&{}\Gamma(\mathrm{ReLU}([\vartheta_{i,0}^T\ \vartheta_{i,1}^T\ ...\ \vartheta_{i,K-1}^T]))\ \mathbf{w}_{total}
\label{}
\end{align}
\end{small}
Finally, we present our temporal hierarchy loss $\mathcal{L}_{th}$ to incorporate the aggregated and stage-wise predictions:
\begin{small}
\begin{align}
\mathcal{L}_{th}=-\mathbb{E}[\vec{c}_i^T log(\hat{\vartheta}_{i,total})]-\sum_{\kappa=0}^{K-1} \mathcal{L}_{\kappa}
\label{}
\end{align}
\end{small}
\subsubsection{Stream Fusion Loss} 
We explore three different mechanisms for fusing the predictions of the SHS and THS streams, to produce the final task prediction logits $f_i^c$. We provide experimental results of each in Section \ref{sec:ablation_study}.

\textbf{Average Fusion.} Here, we treat results of semantic and temporal hierarchies equally, and we backpropagate the same gradient to both streams at training time.
\begin{small}
\begin{align}
f_i^c=0.5(\vartheta_{i,total}+\psi_i^{c})
\label{}
\end{align}
\end{small}
\textbf{Weighted Average Fusion.} Here, the final prediction is a linear combination of streams, whose predictions are weighted by learned weights $\mathbf{w_1},\mathbf{w_2}\in \mathbb{R}^{|\mathbb{C}|\times |\mathbb{C}|}$. .
\begin{small}
\begin{align}
f_i^c=\mathbf{w_1}\vartheta_{i,total}+\mathbf{w_2}\psi_i^{c}
\label{}
\end{align}
\end{small}
\textbf{Task-wise Switching Gates.} Sometimes, wrong predictions of one stream can negatively impact the final fused classification scores  $f_i^c$ at test time. We introduce task-wise switching gates to allow the system enough freedom to learn, for each task independently, to switch between stream predictions. We define switching gate $\alpha=\sigma(w_{\alpha})$ as the sigmoid function $\sigma()$ of learnable parameters $w_{\alpha}\in\mathbb{R}^{|\mathbb{C}|}$. The sigmoid function makes sure our gates stay in the range of 0 to 1. Then, for training and test time, $f_i^c$ is defined as:
\begin{small}
\begin{equation}\label{class}
f_i^c=\left\{\begin{array}{cl}
\alpha\odot\vartheta_{i,total}+(1-\alpha)\odot\psi_i^{c}, &\mathrm{training}\\
\mathcal{H}_{0.5}(\alpha)\odot\vartheta_{i,total}+(1-\mathcal{H}_{0.5}(\alpha))\odot\psi_i^{c},& \mathrm{test}\
\end{array}\right.
\end{equation}
\end{small}
where $\mathcal{H}_{x}()$ denotes the Heaviside step function shifted to x. At test time, given task $\tau$, our final prediction is discretely chosen from the SHS stream if $\alpha_\tau <0.5$ or is selected from the THS stream otherwise. 

In cases of the weighted average fusion and switching gates, our fusion loss $\mathcal{L}_{f}=-\mathbb{E}[\vec{c}_i^T log(\hat{f}_i^c)]$ is added to the previous losses to form our final loss $\mathcal{L}$ with the design parameter $\beta$. Finally we train the whole network end-to-end but stop the gradients of $\mathcal{L}_{f}$ flowing back to the streams to isolate the fusion module from the rest.
\begin{small}
\begin{align}
\mathcal{L}=\mathcal{L}_f+\mathcal{L}_{sh}+\beta\mathcal{L}_{th}
\label{}
\end{align}
\end{small}
\section{Top-Down Action Segmentation}
We now present our top-down segmentation approach.
In the segmentation problem, the goal is to partition a video temporally into a sequence of $S$ action labels $\boldsymbol{\delta}_1^S$ and their corresponding durations $\mathbf{l}_1^S$.
The input in our approach is a video of $T_v$ frames, represented by $\mathbf{x}_1^{T_v}$, as the sequence of per-frame features. Let $\Pi(\tau)$ be the set of all action sequences in the training set given the top-level task $\tau$. Then, grammar $\boldsymbol{\pi}\in\Pi(\tau)$ lists an ordered sequence of $S$ action labels taking place in the video of task $\tau$.
The goal is to identify the most likely sequence of action labels $\overline{\boldsymbol{\delta}}_1^S$ and their durations $\overline{\mathbf{l}}_1^S$ associated with a specific grammar $\boldsymbol{\pi}$:
\begin{small}
\begin{align}
(\overline{\boldsymbol{\delta}}_1^S,\overline{\mathbf{l}}_1^S,\overline{\tau}) ={}& \argmax_{\boldsymbol{\delta}_1^S,\mathbf{l}_1^S,\tau} p(\boldsymbol{\delta}_1^S, \mathbf{l}_1^S,\tau | \mathbf{x}_1^{T_v})  \\
={}& \argmax_{\substack{\boldsymbol{\delta}_1^S\in\Pi(\tau),\mathbf{l}_1^S,\tau}} p(\mathbf{x}_1^{T_v} | \boldsymbol{\delta}_1^S) p(\mathbf{l}_1^S | \boldsymbol{\delta}_1^S) p(\boldsymbol{\delta}_1^S | \tau) p(\tau)
\label{eq:viterbi}
\end{align}
\end{small}
where $p(\mathbf{x}_1^{T_v} | \boldsymbol{\delta}_1^S)$ is modeled by a neural network and the Bayes rule as in \cite{Fine2Coarse,NNViterbi}, and $p(\mathbf{l}_1^S | \boldsymbol{\delta}_1^S)$  is any given duration model, e.g., Poisson\cite{NNViterbi} or DurNet\cite{ghoddoosian2021action}.

Eq.\ref{eq:viterbi} is formulated similarly to the probabilistic model in \cite{NNViterbi}. However, we explicitly integrate the task variable $\tau\in\mathbb{C}$ into this equation, which dictates the choice of the fine-grained action sequence $\boldsymbol{\delta}_1^S$. Specifically we introduce the task model $p(\tau)$ as the probability output of a task classification network. Without loss of generality, we used the output of our two-stream hierarchical network $\hat{f}^c$, so that $p(\tau)=1$ for the predicted task $\tau=\argmax(\hat{f}^c)$ and $p(\tau)=0$ otherwise. In \cite{NNViterbi}, the task is a by-product of the inferred segments ($\overline{\boldsymbol{\delta}}_1^S$,$\overline{\mathbf{l}}_1^S$). In contrast, our proposed approach eliminates all segmentations whose inferred actions do not belong to $\Pi(\tau)$ of the predicted task $\tau$ by setting $p(\boldsymbol{\delta}_1^S | \tau)$ to 0 for those segmentations, and to 1 otherwise.
We follow the Viterbi algorithm in \cite{NNViterbi} to solve Eq. \ref{eq:viterbi}.

\section{Experiments} \label{Experiments}
We compare our method to several existing methods on two popular instructional video datasets, both for task classification and for action segmentation using weakly-labeled videos as training. Further, in ablation studies we evaluate the contribution of each component of our model.

\textbf{Datasets.} 1) \textit{The Breakfast Dataset (BD)}~\cite{Breakfast} consists of around 1.7k untrimmed cooking videos of few seconds to over ten minutes long. There are 48 action labels demonstrating 10 breakfast dishes with a mean  of  4.9  actions  per video, and the evaluation metrics are conventionally calculated over four splits. 2) \textit{The MPII Cooking 2 (C2)}~\cite{Cooking} has training and test subject-wise splits of 201 and 42 long and high quality videos respectively. Particularly, these videos are 1 to 40 mins long adding to 27 hours of data from 29 subjects who prepare 58 different dishes. This dataset offers different challenges compared to the \textit{BD} dataset for two main reasons; First, the annotated 155 objects and 67 actions (verbs) are extremely fine-grained, so that there are on average 51.8 non-background action segments per video. Second, despite the great number of frames in the dataset, the number of samples per class is unbalanced and limited.

\textbf{Metrics.}
We evaluate task classification performance using two metrics: 1) \textit{t-acc} is the standard mean task accuracy over all videos. 2) \textit{t-mAP} denotes the mean Average Precision of task predictions. \textit{mAP} is used in \cite{Cooking} to assign soft class-wise scores to give insight about how far off the wrong predictions are.   
Further, we use four metrics as \cite{TCFPN} to measure the segmentation results: \textit{acc} and \textit{acc-bg} are the frame-level accuracies with and without background frames, while \textit{IoU} and \textit{IoD} define the average non-background intersection over union and detection, respectively.

\textbf{Implementation.} 
We extracted I3D features on the \textit{C2} dataset using TV-L1 optical flow \cite{TV-L1} on a moving window of 32 frames with stride 2, and the pre-computed I3D features of the \textit{BD} dataset were obtained from \cite{fullysupervised5}. 
We noticed that it is not necessary to process the whole video. Instead, we followed the sampling strategy in \cite{WTALC} to maintain the length of the videos in a batch to be less than a pre-defined length $T\approx9$ mins while training. This approach speeds up training, lowers memory demands, and applies temporal augmentation. Also, we divided videos into $K=3$ stages for the THS stream (analysis in Section \ref{THS_ablation}).

Our model is trained with a batch size of 10 using the Adam \cite{adam} optimizer with $10^{-3}$ learning rate and 0.005 weight decay for 20k iterations. For both datasets, we adjust $\lambda$ to 0.9, and $\beta$ is set to 0.25 and 0.01 for the \textit{BD} and \textit{C2} datasets, respectively. The 1D convolutions are done with $F=64$ as the number of kernels, and $L=15$ as their size. $s=8$ and we use a drop-out keep rate of 0.3. The set of verbs and objects are used as our attributes. 

\floatsetup{capposition=top}
\begin{table}
\begin{center}
\footnotesize\setlength{\tabcolsep}{2.3pt}
\caption {\label{classification} Task classification results of state-of-the-art methods on two main datasets. Best results reported out of I3D ($^\dag$) or iDT ($^\ddag$) features (more in supplementary material and \cite{yasser2020evaluating}).* results obtained using the author's source code. \cite{TEC} results for 10 classes.} 
\begin{tabular}{c| l | l l | l l}
 & & \multicolumn{2}{c}{Breakfast (\%)}  &  \multicolumn{2}{c}{Cooking (\%)}    \\ 
\cline{2-6}
Supervision & Models & t-acc & t-mAP & t-acc & t-mAP  \\    \hline

\scriptsize{Full} & \scriptsize{Rohrbach \textit{et al.}\cite{Cooking} } & \,\,\,\,\,- & \,\,\,\,\,- & \,\,\,\,\,- & 57.40\\
\scriptsize{Unsupervised} & \scriptsize{CTE\cite{TEC} } & 31.80$^\ddag$ & \,\,\,\,\,- & \,\,\,\,\,- & \,\,\,\,\,-\\
\hline
{} & \scriptsize{NNViterbi\cite{NNViterbi}$^*$} & 70.98$^\ddag$ & \,\,\,\,\,- & 23.80$^\dag$ & \,\,\,\,\,-\\
{} & \scriptsize{CDFL}\cite{CDFL}$^*$ & 74.86$^\ddag$ & \,\,\,\,\,- & 28.57$^\dag$ & \,\,\,\, -\\
\scriptsize{Weak} & \scriptsize{W-TALC}\cite{WTALC}$^*$ & 76.19$^\dag$ & 80.98$^\dag$ & 33.33$^\dag$ & 43.07$^\dag$\\
{} &\scriptsize{3C-Net}\cite{3C}$^*$ & 75.23$^\dag$ & 80.99$^\dag$ & 30.95$^\dag$ & 46.30$^\dag$\\
{} &\scriptsize{Timeception}\cite{timeception}$^*$ & 76.37$^\dag$ & 80.80$^\dag$ & 21.43$^\dag$ & 25.14$^\dag$\\
{} &\scriptsize{VideoGraph}\cite{videograph}$^*$ & 78.70$^\dag$ & \,\,\,\,\,- & 23.80$^\dag$ & \,\,\,\,\,-\\
\cline{2-6}
{} &\scriptsize{Our Method} & \textbf{80.04} & \textbf{86.36} & \textbf{45.24} & \textbf{54.49}\\
\end{tabular}
\end{center}
\end{table}

\subsection{Comparison to State-of-the-Art Methods}
We used the standard dish labels in both datasets as task labels. All experiments on the \textit{BD} dataset for all models, except the unsupervised CTE \cite{TEC}, were done for 9 tasks after we combined the two dishes of \textit{frying} and \textit{scrambling eggs} as the top-level task of \textit{making eggs}, because both share almost the same set of actions. For CTE, we report the results on the original 10 classes, as given by the authors. We note that CTE is unsupervised and not a direct competitor.

\textbf{Task Classification.}
Table \ref{classification} shows quantitative results on task recognition, for our method as well as other methods that use different types of supervision. Particularly, \cite{CDFL,NNViterbi} are the state-of-the-art open-source weakly-supervised segmentation methods. They implicitly identify the task corresponding to the inferred sequence of actions during inference. Our explicit task modeling significantly outperforms them in accuracy by around 5 to 9 percent on the \textit{BD} dataset, and by 16 to 21 percent in the 58 tasks of the \textit{C2} dataset.

Originally, \cite{3C,WTALC} are the state-of-the-art open-source weakly-supervised methods with specific loss functions to classify and localize sparse action instances in videos. To compare with them, we trained both to classify tasks. Also, \cite{timeception} and \cite{videograph} classify tasks in long videos by training multi-scale temporal convolutions and graph based representations respectively. Both networks make heavy use of memory and suffer from overfitting specifically in the \textit{C2} dataset, where using low-level attributes is key. 
While such direct task modelings under weak supervision prove to be more effective than the implicit classification using fine-grained action segments \cite{CDFL,NNViterbi}, our hierarchical approach outperforms all competitors considerably in all metrics and datasets.
Table \ref{classification} shows our \textit{t-mAP} on the \textit{C2} dataset comes close to the fully-supervised baseline \cite{Cooking}, which is trained on frame-level action ground-truth. Comparison results on 10 classes of the \textit{BD} dataset are in the supp. material.

\textbf{Action Segmentation.}
Table \ref{segmentation} shows results for action segmentation. In our experiments, we combined our two-stream task prediction framework on top of the state-of-the-art weakly-supervised segmentation methods \cite{CDFL,NNViterbi} and achieved new state-of-the art results on both datasets, manifested more vividly in \textit{acc-bg}, because background frames are independent of the task. Therefore, excluding background frames highlights the contribution of the correct task label in segmentation. This consistent improvement in all metrics, while decreasing the inference time by 80-90\% (Table \ref{runtime}), demonstrates the potential of the proposed top-down approach for weakly-supervised segmentation. Moreover, \textit{CDFL+GT} in Table \ref{segmentation} represents CDFL segmentation results constrained by ground-truth task labels, which serves as an upper bound for our proposed top-down model. 

\begin{table}
\begin{center}
\footnotesize\setlength{\tabcolsep}{2.3pt}
\caption {\label{segmentation}Consistent performance gain in weakly-supervised action segmentation following our proposed top-down approach. I3D and iDT features used for experiments on the \textit{C2} and \textit{BD} datasets, respectively. (* as specified in Table \ref{classification}, **: no source code).} 

\begin{tabular}{ l | l l l l | l l l l}
 & \multicolumn{4}{c}{Breakfast (\%)}  &  \multicolumn{4}{c}{Cooking (\%)}    \\ 
\cline{2-5}\cline{6-9}  
Models &  acc & acc-bg & IoU & IoD  & acc &  acc-bg   & IoU & IoD  \\    \hline

\scriptsize{TCFPN}\cite{TCFPN}$^*$ & 38.4 & 38.4 & 24.2 & 40.6 & 26.9 & 30.3 & 9.5 & \textbf{17.0} \\
\scriptsize{D3TW}\cite{D3TW}$^{**}$ & 45.7 & \,\,\, - & \,\,\,- & \,\,\,\,- & \,\,\,\,- & \,\,\,\,\,- & \,\,\,\,- & \,\,\,\,-\\
\scriptsize{DP-DTW}\cite{ActPrototype}$^{**}$ & 50.8 & \,\,\, - & \textbf{35.6} & 45.1 & \,\,\,\,- & \,\,\,\,\,- & \,\,\,\,- & \,\,\,\,-\\
\scriptsize{NNViterbi}\cite{NNViterbi}$^*$ & 43.6 & 42.5 & 27.8 & 39.2 & 23.5 & 21.2 & 7.7 & 10.9\\
\scriptsize{CDFL}\cite{CDFL}$^*$ & 50.2 & 50.4 & 33.5 & 45.6 & 29.9 & 32.2 & 11.0 & 13.8\\
\hline
\scriptsize{NNViterbi+Ours} & 46.2 & 46.1 & 30.2 & 42.2 & 26.9 & 25.0 & 9.6 & 12.7\\
\scriptsize{CDFL+Ours} & \textbf{51.4} & \textbf{52.0} & 34.5 & \textbf{46.7} & \textbf{31.3} & \textbf{34.5} & \textbf{12.8} & 15.6\\
\hline
\textit{\scriptsize{CDFL+GT}} & \textit{59.8} & \textit{63.0} & \textit{41.3} & \textit{55.2} & \textit{35.0} & \textit{39.7} & \textit{14.4} & \textit{17.6}\\
\end{tabular}
\end{center}
\end{table}

\begin{table}
\begin{center}
\footnotesize\setlength{\tabcolsep}{5pt}
\caption {Inference run time (minutes) improvement of state-of-the-art following the proposed top-down approach for segmentation.}\label{runtime} 

\begin{tabular}{ c | c | c}

{Models} &  Breakfast (split 4) & Cooking  \\    \hline

 \scriptsize{NNViterbi}\cite{NNViterbi}  &  100 &  840\\
 \scriptsize{CDFL}\cite{CDFL}  & 144 &  1070 \\
 \hline
 \scriptsize{NNViterbi+Ours}  & 21 &  64 \\
 \scriptsize{CDFL+Ours}  & 25 &  110 \\ 
 
\end{tabular}
\end{center}
\end{table}


\begin{table}[b]
\begin{center}
\footnotesize\setlength{\tabcolsep}{2.3pt}
\caption {Stream-specific ablation study for task classification.\label{stream_results} }

\begin{tabular}{ c | c c | c c}
 & \multicolumn{2}{c}{Breakfast (\%)}  &  \multicolumn{2}{c}{Cooking (\%)}    \\ 
\cline{2-5}
Stream & t-acc & t-mAP & t-acc & t-mAP  \\    \hline

\scriptsize{Semantic hierarchy} & 73.1 & 77.2 & 42.9 & 52.6\\
\scriptsize{Temporal hierarchy-stage 1} & 62.7 & -& 16.7 & -\\
\scriptsize{Temporal hierarchy-stage 2} & 68.9 & - & 28.6 & -\\
\scriptsize{Temporal hierarchy-stage 3} & 64.7 & - & 23.8 & -\\
\scriptsize{Temporal hierarchy-aggregated} & 80.0 & 86.4 & 31.0 & 45.2\\
\scriptsize{Two streams fused} & \textbf{80.0} & \textbf{86.4} & \textbf{45.2} & \textbf{54.5}\\
\end{tabular}
\end{center}
\end{table}


\subsection{Analysis and Ablation Study in Task Modeling}\label{sec:ablation_study}
\textbf{Stream-Specific Results.} We evaluated the contribution of the SHS and THS streams separately in Table \ref{stream_results}. The SHS stream is more effective on the \textit{C2} dataset because of two main reasons: First, the average number of videos per task (3.4) is low compared to that of videos per attribute (28.4), so any direct way of task modeling is prone to overfitting. Second, the large number of attributes per task allows the learning of a discriminative attribute-to-task mapping. Meanwhile in the THS stream, despite the weak classification power of the stage-specific classifiers, our hierarchical modeling is able to aggregate stage-wise predictions effectively and produce significantly superior results. This shows that different stages provide complimentary information. Note that the THS stream alone achieves state-of-the-art on the \textit{BD} dataset with only task label supervision.

\textbf{Semantic Hierarchy Ablation.} 
As shown in Table \ref{shs_ablation}a, removing the attributes from the semantic hierarchy loss (Eq.\ref{eq:sh_loss}), and directly classifying tasks from the feature encoding $\mathbb{\phi}(\mathbf{x})$, leads to around $12\%$ drop in \textit{t-acc} on the \textit{C2} dataset. Simply sharing low-level attributes among tasks is beneficial, and using the TF-IDF weights led to an additional $7\%$ difference in \textit{t-acc} (Table \ref{shs_ablation}b).

\begin{table}
\begin{center}
\setlength{\tabcolsep}{2.3pt}
\caption {The effect of sharing attributes (a) and using TF-IDF weights (b) in the SHS stream (Cooking dataset).\label{shs_ablation} }

\begin{tabular}{ cc }   
(a) & (b) \\  
\footnotesize
        \begin{tabular}{ c | c c |}
{Setting} &  t-acc & t-mAP  \\    \hline
 \scriptsize{Without attributes}  &  33.3 &  45.4\\
 \scriptsize{With attributes}  & \textbf{45.2} &  \textbf{54.5} \\
\end{tabular} &  
\footnotesize
\begin{tabular}{| c | c c }
{Setting} &  t-acc & t-mAP  \\    \hline
 \scriptsize{Without TF-IDF}  &  38.1 &  49.4\\
 \scriptsize{With TF-IDF}  & \textbf{45.2} &  \textbf{54.5} \\
\end{tabular} \\
\end{tabular}
\end{center}
\end{table}

\textbf{Temporal Hierarchy Analysis.}  \label{THS_ablation}
Modeling tasks as a temporal hierarchy of multiple stages improves the performance compared to the single-stage approach. Furthermore, as indicated by Table \ref{K_ablation}, such an approach is not sensitive to the number of stages ($K>1$) in the hierarchy. This concludes that these stages provide complimentary information for the stage-aggregation function regardless of their exact positioning or duration in the video.

\floatsetup{captionskip=2pt} 
\begin{table}[t]
\setcellgapes{0pt}
  \begin{floatrow}[4]
    \makegapedcells
    \ttabbox%
    {\footnotesize
    \setlength\tabcolsep{4pt} 
\begin{tabular}{ c | c c |}
{$\#$ Stages} &  t-acc & t-mAP  \\    \hline
 \scriptsize{1}  &  74.2 &  82.1\\
 \scriptsize{2}  & 79.6 &  86.4 \\
  \scriptsize{3}  & 80.0 &  86.4 \\
   \scriptsize{4}  & 80.0 &  86.4 \\
    \scriptsize{5}  & 79.2 &  85.8 \\
\end{tabular}}
    {\caption{Evaluation of our model under different choices of $K$ on the \textit{BD}.}
      \label{K_ablation}}
    \hfill%
    \ttabbox%
    {\footnotesize
    \setlength\tabcolsep{2pt} 
\begin{tabular}{| c | c c c}
{Fusion Type} &  t-acc & t-mAP  \\    \hline
 \scriptsize{Average}  &  77.2 &  84.2 & {}\\
 \scriptsize{Weighted Average}  & 78.4 &  84.9 & {}\\
  \scriptsize{Switching Gate}  & \textbf{80.0} &  \textbf{86.4} & {}\\
\end{tabular}}
    {\caption{Comparison between different fusion mechanisms on the \textit{BD}.}
      \label{fusion}}
  \end{floatrow}
\end{table}%

\floatsetup{capposition=bottom}
\begin{figure*} [t]
    \includegraphics[width=1\linewidth]{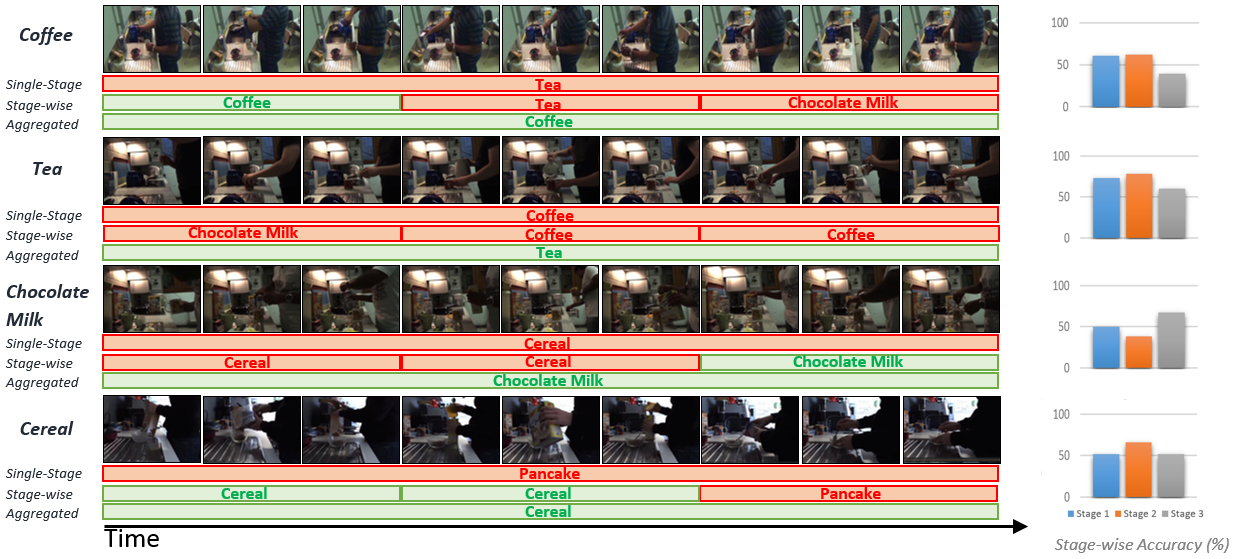}
    \caption{Qualitative task classification results of our THS stream on sample challenging videos from the \textit{BD} dataset. For each video, we show the ground-truth task label, predictions from the single-stage baseline, and the stage-wise and stage-aggregated recognition result. Also, stage-wise accuracy for each of the four tasks is presented on the right side. In the first video, the subject sequentially takes a cup, pours coffee and milk, adds sugar and stirs coffee. In the second video, the subject takes a cup, adds teabag, pours hot water, spoons sugar and stirs. In the third video, the subject takes a cup, pours milk, spoons choc. powder, adds sugar and stirs. In the fourth video, the subject pours cereal, spoons choc. powder, pours milk, and then stirs. Red and green boxes denote wrong and correct predictions, respectively.}
    \label{qualitative}
\end{figure*}

\textbf{Comparison of the Stream Fusion Mechanisms.}
Table \ref{fusion} compares different mechanisms to fuse the predictions of SHS and THS streams. Specifically, the Task-wise Switching Gates are trained to identify the stronger stream per task and perform best, while the vanilla and weighted averaging compromise between both streams and produce sub-optimal results. In the \textit{BD} dataset these gates propagate the results of the THS stream, whereas in the \textit{C2} dataset they switch between both for different tasks and combine predictions (see row 1, 5 and 6 of Table \ref{stream_results}).

\subsection{Qualitative Results}
Fig.\ref{qualitative} compares results of the THS stream with the single-stage baseline on four challenging videos from the \textit{BD} dataset. Due to similar-looking actions shared between tasks, the single stage baseline misclassifies the video task, whereas our THS stream outputs the correct task under different stage-wise settings. Specifically, in the first and third videos, our model classifies the task correctly although only one of the stage predictions is correct. For example, according to the stage-wise accuracy of the task \textit{making chocolate milk} in Fig.\ref{qualitative}, the last stage of this task is the most discriminative one. Thus, our model learns to put more weight on the predictions of this stage, which compensates for the first two stages outputting the wrong class of \textit{making cereal}.

The two tasks of \textit{making coffee} and \textit{making tea} share similar-looking actions in the first and second videos, so analyzing the entire video in one step produces wrong predictions for both cases. The second video, in particular, provides an interesting case where similar visuals of the action \textit{taking cup} between tasks of \textit{making chocolate milk} and \textit{making tea} led to confusion of the first stage. Also, the later two stages mistakenly predicted the task of \textit{making coffee}, because the two actions of \textit{pouring} and \textit{stirring} are shared between both tasks of \textit{making coffee} and \textit{making tea}. Although all three stage-wise predictions are wrong, the aggregated result of those stages is correct. This shows that the proposed hierarchical model not only considers the predicted class of each stage, but also learns the relationship between stages and their fine-grained prediction scores.  

In a given stage, the short discriminative part may be dominated by the longer ambiguous section. For example, the final stage of the last video depicts how \textit{stirring}  while  occluding  the  bowl  dominates the shorter and more discriminative action of \textit{pouring milk}. This effectively resembles the appearance and motion of \textit{flipping a pancake by spatula}, but the complementary information of the first two stages eventually results in the correct aggregated recognition.
\section{Conclusion}
We have introduced a two-stream framework, that exploits semantic and temporal hierarchies to recognize tasks in weakly-labeled instructional videos. We have also proposed a novel top-down segmentation approach, where the predicted task constrains the fine-grained action labels. We report experimental results on two public datasets. Our two-stream task recognition method outperforms existing methods. Similarly, our top-down segmentation approach improves the accuracy of existing state-of-the-art methods, while simultaneously improving runtime by 80-90\%.

{\small
\bibliographystyle{ieee_fullname}
\bibliography{egbib}
}

\newpage
\clearpage

\section{Supplementary Material}
In this supplementary material, we show comparisons of I3D \cite{I3D} and iDT \cite{iDT} features in task recognition on two datasets, and present comparison results on the original 10 classes of the Breakfast dataset \cite{Breakfast}. We also provide a glossary of terms and a table of symbols we use in the paper.
\subsection{I3D and iDT Feature Comparison in Task Recognition of Weakly-Labeled Videos}
In this section, we compare  I3D and iDT features for the purpose of task recognition in weakly-labeled instructional videos. Specifically, we present results of existing models using I3D (Table \ref{I3D}) and iDT (Table \ref{iDT}) features on the MPII Cooking 2 dataset \cite{Cooking} as well as the first split of the Breakfast dataset \cite{Breakfast}. 

We used the Fisher vectors of iDT features as in \cite{Breakfast,actionset}. The Fisher vectors for each frame are extracted over a sliding window of 20 frames. They are first projected to a 64-dimensional space by PCA, and then normalized along each dimension.
Also, we extracted the I3D features of the Cooking 2 dataset using TV-L1 optical flow \cite{TV-L1} on a moving window of 32 frames with a stride of 2, and the pre-computed I3D features of the Breakfast dataset were obtained from \cite{fullysupervised5}.
Furthermore, we applied PCA to the extracted I3D features to reduce the dimensionality of RGB and optical flow channels from 1024 to 128. We fed the same features to all competitors except \cite{rhyrnn} in Table \ref{BD10} whose code is not publicly available, so we compare with their reported result on ResNet101 \cite{ResNet101} features.

In the Cooking 2 dataset, we train the models on the training split and test on the test split. However as \cite{CDFL} and \cite{NNViterbi} take a long time to train and infer the segments, in Tables \ref{I3D} and \ref{iDT}, we only use the first split of the Breakfast dataset to evaluate the difference in performance of all models when using I3D and iDT features as input. Note that the reported task recognition results on the Breakfast dataset in the paper are the average of all four splits using the best case for each method. 

 Explicit task classification methods, e.g., ours, W-TALC \cite{WTALC} and 3C-Net \cite{3C}, consistently perform better with I3D features on both datasets, whereas the bottom-up inference of tasks in NNViterbi \cite{NNViterbi} and CDFL \cite{CDFL} produces mixed result. In particular, the performance of \cite{CDFL} and \cite{NNViterbi} on the Breakfast dataset considerably improves upon using iDT features. Overall, the more significant presence of object information in I3D features helps to classify top-level tasks more accurately, while detecting fine-grained actions seems to be less affected by such appearance information. 
\begin{table}[t]
\begin{center}
\setlength{\tabcolsep}{2.3pt}
\caption {\label{I3D} Task classification results of state-of-the-art methods using I3D features on the Cooking 2 dataset and the first split of the Breakfast dataset. (* results obtained using the author's source code).}
\begin{tabular}{ l | l l | l l}
 & \multicolumn{2}{c}{Breakfast (1st split) (\%)}  &  \multicolumn{2}{c}{Cooking (\%)}    \\ 
\cline{2-5}
 Models & t-acc & t-mAP & t-acc & t-mAP  \\    
\hline
NNViterbi\cite{NNViterbi}$^*$ & 57.14 & \,\,\,\,\,- & 23.80 & \,\,\,\,-\\
CDFL\cite{CDFL}$^*$ & 66.26 & \,\,\,\,\,- & 28.57 & \,\,\,\,- \\
W-TALC\cite{WTALC}$^*$ & 75.79 & 78.96 & 33.33& 43.07\\
3C-Net\cite{3C}$^*$ &  75.39 & 78.50 & 30.95 & 46.30\\
Timeception\cite{timeception}$^*$ &  79.50 & 82.53 & 21.43 & 25.14\\
VideoGraph\cite{videograph}$^*$ &  80.06 & \,\,\,\,\,- & 23.80 & \,\,\,\,-\\
\hline
Our Method & 81.74 & 88.30 & 45.24 & 54.49\\
\end{tabular}
\end{center}
\end{table}

\begin{table}[t]
\begin{center}
\setlength{\tabcolsep}{2.3pt}
\caption {\label{iDT} Task classification results of state-of-the-art methods using iDT features on the Cooking 2 dataset and the first split of the Breakfast dataset. (* results obtained using the author's source code).}
\begin{tabular}{ l | l l | l l}
 & \multicolumn{2}{c}{Breakfast (1st split) (\%)}  &  \multicolumn{2}{c}{Cooking (\%)}    \\ 
\cline{2-5}
 Models & t-acc & t-mAP & t-acc & t-mAP  \\    
\hline
NNViterbi\cite{NNViterbi}$^*$ & 71.03 & \,\,\,\,\,- & 16.66 & \,\,\,\,-\\
CDFL\cite{CDFL}$^*$ & 77.38 & \,\,\,\,\,- & 21.42 & \,\,\,\,-\\
W-TALC\cite{WTALC}$^*$ & 53.17 & 54.96 & 19.04 & 25.85\\
3C-Net\cite{3C}$^*$ & 56.74 & 60.36 & 14.28 & 27.38\\
Timeception\cite{timeception}$^*$ &  65.87 & 71.73 & 9.52 & 14.36\\
VideoGraph\cite{videograph}$^*$ &  58.93 & \,\,\,\,\,- & 14.28 & \,\,\,\,-\\
\hline
Our Method & 60.31 & 61.72 & 23.80 & 27.66\\
\end{tabular}
\end{center}
\end{table}

\subsection{Task Classification Results on 10 Classes of the Breakfast Dataset}

Timception \cite{timeception}, VideoGraph \cite{videograph} and RhyRNN \cite{rhyrnn} are the latest state-of-the-art methods to classify tasks in minutes-long videos and are the closest competitors to our work. We compared the standard four fold cross validated results of  Timeception and VideoGraph over 9 classes of the Breakfast dataset in Table 1 of the paper, however, we could not compare our method to RhyRNN because the source code of RhyRNN is not publicly available to adjust that model to our evaluation settings. Hence, in Table \ref{BD10}, we present comparison results of our method with the reported accuracy of this method and different versions of other models over the original 10 classes of the Breakfast dataset. For a direct comparison with RhyRNN , we show results on the first split as reported in RhyRNN. 

Furthermore, Table \ref{BD10} shows the original reported results of Timeception and VideoGraph, which are lower than our re-implemented versions in both cases. Contrary to the standard splitting rule of the Breakfast dataset, both works have used the last 0.15\% of subjects in the dataset (8 subjects) to test their performance. Our result on this split significantly outperforms previous methods (Table \ref{BD10}). \cite{timeception} and \cite{videograph} also use the output before the last average pooling layer (\textit{pre pooling}) in the I3D network as features, unlike us, where we use the features after the pooling layer (\textit{post pooling}). The results in Table \ref{BD10} suggest the superiority of the latter, because the lower dimension after pooling allows each network to be given more features as input, which increases their input temporal range.

Interestingly, the task accuracy for most models, including ours, hardly drops upon evaluation on 10 classes and our method is still superior than different versions of state-of-the-art.
\begin{table}[t]
\begin{center}
\setlength{\tabcolsep}{2.3pt}
\caption {\label{BD10} Task classification results (t-acc) of state-of-the-art methods on the Breakfast dataset for 10 classes. (* results re-implemented using the author's source code).}
\begin{tabular}{ l | l | c | c}
 Models & t-acc & Feature & Test Split  \\    
\hline
Timeception\cite{timeception} &  71.3 &  3D-ResNet \cite{3DResNet} & Last 8 subjects\\
Timeception\cite{timeception} &  69.3  &  I3D (\textit{pre pooling}) & Last 8 subjects\\
Timeception\cite{timeception}$^*$ &  76.6 &  I3D (\textit{post pooling}) & Split 1\\
VideoGraph\cite{videograph} &  69.5 &  I3D (\textit{pre pooling}) & Last 8 subjects\\
VideoGraph\cite{videograph}$^*$ &  79.9 &  I3D (\textit{post pooling}) & Split 1\\
RhyRNN\cite{rhyrnn} &  44.3 &  ResNet101 \cite{ResNet101} & Split 1\\
\hline
Our Method & 81.5 & I3D (\textit{post pooling}) & Split 1\\
Our Method & 85.2 & I3D (\textit{post pooling}) & Last 8 subjects\\
\end{tabular}
\end{center}
\end{table}

\subsection{Glossary of Terms and Symbols}
As there are similar terms and many symbols used in the paper, here, we provide specific definitions of terms (Table \ref{terms}) and symbols (Table \ref{symbols}) for readers to refer to.
   
\begin{table}[h]
\centering
\footnotesize
\caption {Definitions of technical terms used in the paper.\label{terms} }
\begin{tabular}{|p{1.8cm}  | p{5cm} |}
\hline
\textbf{Term} & \textbf{Definition}  \\    \hline

\scriptsize{Action} & Lower level actions happening in the form of segment sequence in instructional videos. \\
\hline
\scriptsize{Action alignment} & Partitioning the video into sequence of action segments given a sequence of action labels. \\
\hline
\scriptsize{Action detection} & Classify and localize occurrences of, typically, a single action in the video among considerable background frames.\\
\hline
\scriptsize{Action segmentation} & Partitioning the video into sequence of action segments. \\
\hline
\scriptsize{Attribute} & Set of actions or the set of verb/object components of actions.\\
\hline
\scriptsize{Fully-supervised classification} & Task classification using frame-level and video-level labels.\\
\hline
\scriptsize{Instructional videos} & Videos with a top-level task and a sequence of fine-grained actions to carry out the underlying task. \\
\hline
\scriptsize{Task} & The single top-level composite activity present in the video. \\
\hline
\scriptsize{Task recognition} & Classifying the top-level task in long instructional videos. \\
\hline
\scriptsize{Weakly-labeled videos} & Videos with no frame-level annotations. In our case, only sequence of video-level action labels is available. \\
\hline
\scriptsize{Weakly-supervised classification} & Task classification without access to frame-level annotation. We use the term ``weak'' to distinguish from fully-supervised methods.\\
\hline

\end{tabular}
\end{table}

\newpage
\begin{table}[h]
\begin{center}
\footnotesize\setlength{\tabcolsep}{2.3pt}
\caption {Definitions of symbols used in the paper.\label{symbols} }

\begin{tabular}{| l | l |}
\hline
\textbf{Symbol} &\textbf{ Definition}  \\    \hline

\scriptsize{$\mathbb{A}$} & The set of all attributes \\
\hline
\scriptsize{$a_{j}$} & Attribute j \\
\hline
\scriptsize{$a_{i,j}$} & Attribute j of video i \\
\hline
\scriptsize{$\mathbb{A}_i$} & The set of attributes in video i \\
\hline
\scriptsize{$\vec{a}_{i}$} & Muiltihot ground-truth attribute vector of video i \\
\hline
\scriptsize{$\vec{a}_{i}^w$} & TF-IDF weighted ground-truth attribute vector of video i \\
\hline
\scriptsize{$A^TB$} & Matrix multiplication of A transposed and B \\
\hline
\scriptsize{$a \cdot b$} &Scalar multiplication of a and b \\
\hline
\scriptsize{$\beta$} & Importance factor of  $\mathcal{L}_{th}$ in the total loss\\
\hline
\scriptsize{$\mathbb{C}$} & The set of all tasks \\
\hline
\scriptsize{$c_{i}$} & Task label for video i \\
\hline
 \scriptsize{$\vec{c}_i$} & One-hot task ground-truth vector of video i \\
 \hline
\scriptsize{$d$} & Stage duration \\
\hline
\scriptsize{$F$} & Dimension of the feature encoding $\phi(\mathbf{x})$ \\
\hline
\scriptsize{$f_i^c$} & Final fused classification logits \\
\hline
\scriptsize{$g(\mathbf{x})$} & The fully connected layer to produce encoding $\phi(\mathbf{x})$ \\
\hline
\scriptsize{$\mathcal{H}_{x}(\ )$} & Heaviside step function shifted to x \\
\hline
\scriptsize{$h_{\kappa}$} & Feature summary of stage $\kappa$ \\
\hline
\scriptsize{K} & Number of stages in the THS stream \\
\hline
\scriptsize{$\mathbf{k}_{\phi}$} & Temporal convolution kernels to produce $\phi(\mathbf{x})$ \\
\hline
\scriptsize{$k_i$} & Number of selected frames of video i from the $\mathrm{top_k}$ operation \\
\hline
\scriptsize{L} & Kernel length of $\mathbf{k}_{\phi}$ \\
\hline
\scriptsize{$\boldsymbol{l}_1^S$} & Sequence of S action durations in a video \\
\hline
\scriptsize{$\mathcal{L}_{sh}$} & Loss function for the SHS stream \\
\hline
\scriptsize{$\mathcal{L}_{th}$} & Loss function for the THS stream \\
\hline
\scriptsize{$\mathcal{L}_{f}$} & Loss function of the fused streams \\
\hline
\scriptsize{$M_i$} & Number of attributes in video i \\
\hline
\scriptsize{$\mathbf{M}_{tfidf}$} & TF-IDF mask \\
\hline
\scriptsize{$\mathcal{M}_x^a$} & Mapping function from features to attributes \\
\hline
\scriptsize{$\mathcal{M}_a^c$} & Mapping function from attributes to tasks \\
\hline
\scriptsize{N} & Number of videos in the training set/batch \\
\hline
\scriptsize{$S$} & Number of segments in a video \\
\hline
\scriptsize{$\mathfrak{s}[\ ]$} & Softmax operation \\
\hline
\scriptsize{$\mathcal{S}_{\kappa}$} & Classifier for stage $\kappa$ in the THS stream \\
\hline
 \scriptsize{$s$} & The parameter used in the $\mathrm{top_k}$ operation \\
 \hline
\scriptsize{$T_i$} & Number of frames in video i \\
\hline
\scriptsize{$\mathcal{T}$} & Stage aggregation function in the THS stream \\
\hline
\scriptsize{$\tau$} & Task variable \\
\hline
\scriptsize{$\vartheta_{\kappa}$} & Task prediction logits of stage $\kappa$ \\
\hline
\scriptsize{$\vartheta_{total}$} & Stage-aggregated task prediction logits \\
\hline
\scriptsize{$\mathbb{V}$} & Set/Batch of training videos \\
\hline
\scriptsize{$\mathbf{v}_i$} & Video $i$ \\
\hline
\scriptsize{$\mathbf{W}_{tfidf}$} & TF-IDF weights \\
\hline

\scriptsize{$\mathbf{x}_i$} & Input feature vector for video i \\
\hline
\scriptsize{$\phi(\mathbf{x})$} & Learned video feature encoding  \\
\hline
\scriptsize{$\psi_i^a$} & Attribute score vector of video i in the SHS stream \\
\hline
\scriptsize{$\psi_i^c$} & Task score vector of video i in the SHS stream \\
\hline
\scriptsize{$\Psi_i^a$} & T-CAM of video i \\
\hline
\scriptsize{$\boldsymbol{\delta}_1^S$} & Sequence of S action labels in a video \\
\hline
\scriptsize{$\lambda$} & Design parameter in  $\mathcal{L}_{sh}$\\
\hline
\scriptsize{$\Pi(\tau)$} & Set of all action sequences in the training set given task $\tau$ \\
\hline
\scriptsize{$\sigma (\ )$} & Sigmoid operation \\
\hline
\scriptsize{$\Gamma(\ )$} & Stage-wise drop out in the stage aggregation function \\
\hline
\scriptsize{$\mathbbm{1}(\ )$} & Indicator function \\
\hline
\scriptsize{$\odot$} & Element-wise product operation \\
\hline
\end{tabular}
\end{center}
\end{table}


\pagebreak

\end{document}